\documentclass{article}

    \PassOptionsToPackage{numbers}{natbib}
 \usepackage[preprint]{neurips_2026}

\usepackage[utf8]{inputenc} 
\usepackage[T1]{fontenc}    
\usepackage{hyperref}       
\usepackage{url}            
\usepackage{booktabs}       
\usepackage{amsfonts}       
\usepackage{nicefrac}       
\usepackage{microtype}      
\usepackage{xcolor}         

\usepackage{amsmath}
\usepackage{enumitem}
\usepackage[flushleft]{threeparttable}
\usepackage{colortbl}
\usepackage[table]{xcolor}
\usepackage{graphicx}
\usepackage{tabularx}
\usepackage{array}
\usepackage{subcaption}
\usepackage{caption}
\definecolor{softred}{RGB}{255, 180, 180}
\definecolor{softgreen}{RGB}{180, 225, 180}
\usepackage{fontawesome5}

\usepackage{tcolorbox}
\usepackage{listings}
\tcbuselibrary{listings}
\DeclareCaptionType{listing}[Listing][List of Listings]
\newcommand{\faHuggingFace}{%
  \raisebox{-0.15em}{\includegraphics[height=1em]{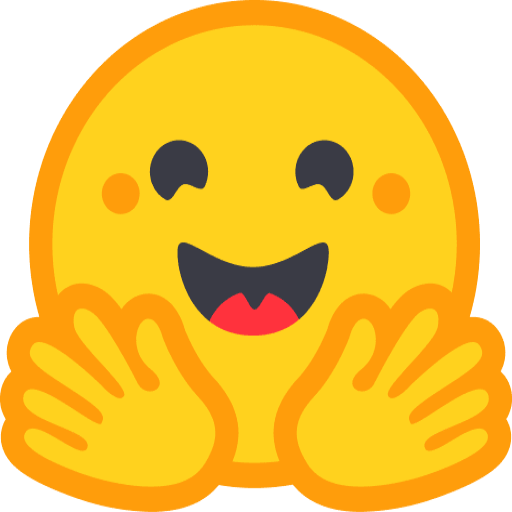}}%
}
\lstdefinestyle{promptstyle}{
    basicstyle=\ttfamily\small,
    breaklines=true,
    columns=fullflexible,
    keepspaces=true,
    frame=none,
    showstringspaces=false
}

\title{On the Fragility of Self-Improving Agents: 
\\Variance, Task Order, and Underspecification}

\author{%
Qinyuan Ye\quad Yu Li\quad Yada Pruksachatkun\quad Jiaxin Zhang\quad Chien-Sheng Wu
\\Salesforce AI Research
}

\begin{document}

\maketitle

\begin{abstract}
Memory-based self-improving agents—those that learn from an online stream of tasks and improve over time by maintaining a textual memory bank—have shown great promise in recent literature. However, the reliability aspects of these methods have been critically overlooked.
In this work, we conduct a comprehensive re-evaluation of two memory-based methods, broadening the scope of evaluation along two axes: (1) including multiple self-improving runs to quantify \textit{variance}, and (2) randomly shuffling the tasks to investigate the effect of \textit{task order}. Through these experiments, we make two observations that expose the fragility of current methods:
First, agent evaluation is inherently noisy in complex environments and on multi-step tasks, and stacking a self-improving loop on top can further amplify this noise. Empirically, when these methods are applied, we observe the variance across runs increase in 71\% of cases, and the gap between the best and worst runs of the same experiment can reach up to 10 percentage points.
Second, the agent's improvement is highly dependent on task order. Prior works often adopt default orderings that impose an implicit curriculum, acting as a hidden prerequisite for success. In contrast, when evaluated under a shuffled task order, agent performance degrades (-4.5\%) instead of exhibiting the expected improvement (+1.5\%).

To better understand this fragility, we manually examine the agents' memory and hypothesize that task and environment \textit{underspecification} contribute to this fragility. Without clear specifications, agents generate plausible yet inapplicable memories (\textit{e.g.}, recommending API usage in a browser-only environment) that may distract the agent from feasible strategies. We validate this hypothesis by incorporating information that enables better specification, such as detailed rubrics and environment feedback, into the memory construction process. While this added information partially closes the performance degradation in previous experiments, significant gaps still remain, suggesting that other uncharacterized factors contribute to this fragility. 
Looking ahead, our work advocates for more rigorous evaluation protocols for self-improving agents by reporting results across multiple runs and stress-testing them under realistic, challenging conditions. Moreover, our findings on underspecification call for systems and interfaces that enable effective human oversight, preventing agents from failing in unforeseeable ways.\footnote{\faGithub\ \href{https://github.com/SalesforceAIResearch/self-improve-fragility}{\texttt{SalesforceAIResearch/self-improve-fragility}} \ \faHuggingFace\ \href{https://huggingface.co/datasets/Salesforce/self-improve-fragility}{\texttt{Salesforce/self-improve-fragility}}}
\end{abstract}

\newtcblisting{promptbox}{
    colback=gray!5,
    colframe=gray!50,
    listing only,
    boxrule=0.5pt,
    arc=2mm,
    left=2mm,
    right=2mm,
    top=1mm,
    bottom=1mm,
    listing options={style=promptstyle}
}

\section{Introduction}
The development of self-improving agents \cite{schmidhuber2006goedelmachinesselfreferentialuniversal,zelikman2022star,huang-etal-2023-large}—systems designed to learn from past experiences and autonomously refine their performance over time—has emerged as a highly promising research direction. If they realize this potential, they could not only better automate routine workflows and create significant economic impact, but also enable highly adaptable systems and facilitate open-ended research discovery. While recent works show progress in this direction \cite{zhuge2024gptswarm,hu2025automated,zhou2026selfchallenging,fang-etal-2025-webevolver,murty2025nnetscape,zheng2025skillweaverwebagentsselfimprove}, little focus has been placed on stress-testing the reliability of such systems \cite{rabanser2026scienceaiagentreliability}. 
In practical settings, such as enterprise deployments, there is often minimal tolerance for error; an initial failure risks losing user trust. 
Moreover, initial mistakes can cascade silently over the long term, causing irreversible or prohibitively costly impacts if adopted in high-stakes domains.
Before we can confidently deploy these systems, we must critically evaluate their reliability under realistic, challenging conditions.

This work begins with a comprehensive re-evaluation of two representative memory-based self-improving systems—Agent Workflow Memory \cite{wang2025agent} and ReasoningBank \cite{ouyang2025reasoningbankscalingagentselfevolving}—across three realistic web-browsing benchmarks: WebArena \cite{zhou2024webarena}, VisualWebArena \cite{koh-etal-2024-visualwebarena}, and SCUBA \cite{dai2025scubasalesforcecomputeruse}. In this evaluation, we establish a stronger baseline by upgrading both the underlying language model and the agent harness. Furthermore, we broaden the scope of evaluation along two axes: conducting multiple self-improving runs of the same experimental setting to quantify variance, and randomly shuffling the tasks to investigate the effect of task order. Through this extended re-evaluation, we identify two critical reliability issues.

First, we show that the evaluation variance can be concerningly large, an observation largely hidden in prior works, as web agent and self-improving agent evaluations typically report only single-run results. 
We show that the no-memory, non-self-improving baseline agent already exhibits significant variance across runs, and that incorporating a self-improving loop can further amplify the variance.
For example, on the WebArena GitLab subset (comprising 180 tasks), the absolute performance gap between the best and worst baseline runs is 4.4\%, whereas applying ReasoningBank enlarges this gap to 7.8\%. These observations call the reliability of single-run evaluation practices into question.

Second, we demonstrate that agent improvement is highly sensitive to task order.
The default task order in prior works imposes an implicit, easy-to-hard curriculum, which acts as a hidden prerequisite for these self-improving methods to succeed.
Under the default order, the agent using ReasoningBank achieves an average performance gain of 1.5\%; however, when evaluated under randomly shuffled task orders, the agent exhibits a performance degradation of 4.5\%. 
This dependency introduces a pressing concern, as real-world applications cannot assume a neatly ordered stream of user requests.

To better understand this fragility, we manually investigate the memories written by the agents, and identify environment and task underspecification during memory generation as potential drivers for the large variance and degraded performance. 
For example, when the memory construction model is unaware that the environment is browser-only and does not support code execution, it generates suggestions to use APIs—a strategy that is plausible but unexecutable.
Regarding task underspecification, we show that ambiguous task queries can give rise to misunderstanding, overthinking, and the construction of erroneous memories that propagate to subsequent tasks.

To mitigate these issues, we consider incorporating additional context—such as environment feedback and task rubrics—into the memory construction step, and we further revise the prompt with clearer environment specifications. These interventions close 31\% of the performance degradation observed in the shuffled task order settings. While this demonstrates the utility of such additional information, the remaining performance gap suggests that other uncharacterized factors contribute to this fragility, highlighting the need for future work to ensure the reliability of self-improving agents.

Beyond these empirical results, our work calls for research into rigorous evaluation methodologies for self-improving AI systems, and we provide practical recommendations in this direction. Furthermore, our analysis of underspecification motivates the development of effective human intervention interfaces, where the ``wrong lessons'' learned by these agents can be identified and corrected timely.

\section{Background: Memory-Based Self-Improving Agents}

In this section, we first revisit the problem setting of memory-based self-improving agents. We then provide a brief overview of two representative methods that we use in our analysis.

\begin{figure}[t]
    \centering
    \includegraphics[width=1\linewidth]{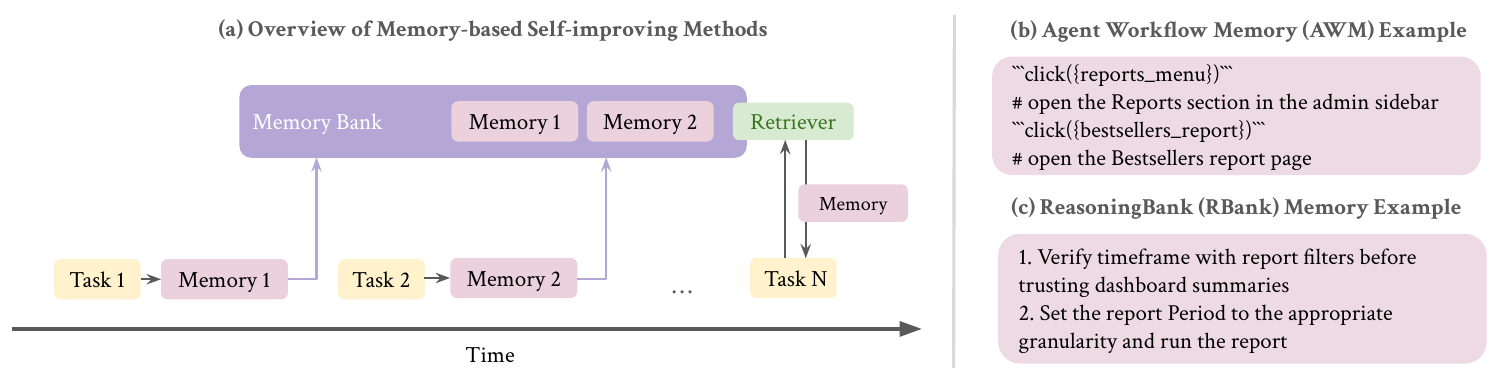}
    \caption{\textbf{(a) Overview of Memory-based Self-improving Methods.} In this setting, agents will perform a sequence of tasks and generate memories at the end of each tasks. The memories will be saved to a memory bank and will be retrieved for performing future tasks. \textbf{(b)(c) Example memories.} We consider two established methods (AWM \cite{wang2025agent} and RBank \cite{ouyang2025reasoningbankscalingagentselfevolving}) in our evaluation.}
    \label{fig:overview}
\end{figure}

\subsection{Problem Setting}
We follow the setting described in prior literature \citep{wang2025agent,wang2025inducing,ouyang2025reasoningbankscalingagentselfevolving}. The agent is required to complete an online stream of $N$ tasks, $\mathcal{Q}=\{q_0, q_1, ..., q_N\}$ and maintain a textual memory $\mathcal{M}$. For each task $q_i$, the agent, backed by a language model $L$, is required to generate an action $a_t\in\mathcal{A}$ from past observations ($o_{0:t-1}$) and actions ($a_{0:t-1}$) at each timestamp $t$, \textit{i.e.}, $a_i\leftarrow\pi_{L}(o_{0:t-1}, a_{0:t-1}; \mathcal{M}, \mathcal{A})$. 
At the end of each task, the agent receives a reward $r\in[0,1]$, where a reward of $1$ indicates that the agent ``passes'' the task. 
A memory construction module $C$ will be instructed to update the memory $\mathcal{M}$ based on the task trajectory and optionally the reward.
See Fig.~\ref{fig:overview}(a) for an illustration. 

\subsection{Representative Approaches}
\label{ssec:representative-approaches}
We consider two representative approaches in our evaluation, and we describe them below.

\begin{itemize}[leftmargin=2em, topsep=0pt, parsep=0pt]
    \item \textbf{Agent Workflow Memory (AWM)} \citep{wang2025agent}. 
    At the end of each task, the memory module is instructed to summarize reusable workflows from successful trajectories (Fig.~\ref{fig:overview}(b)). Following the original implementation, each new workflow is checked for duplication against existing workflows, and all workflows are included in the agent’s context when performing subsequent tasks.
    \item \textbf{ReasoningBank (RBank)} \citep{ouyang2025reasoningbankscalingagentselfevolving}. In this more recent approach, memory items were not limited to structured, step-by-step workflows, but more generic reasoning traces and insights that could be helpful for future similar tasks (Fig.~\ref{fig:overview}(c)). Following the original implementation, memory items are generated from both successful and failed trajectories, and a retriever is used to select the most relevant past memories for subsequent tasks.
\end{itemize}

\section{Exposing Fragility with Broadened Re-evaluation}
\label{sec:exposing-fragility}

\subsection{Experiment Setting}
\label{ssec:experiments}
\paragraph{Benchmarks.} In this work, we focus our evaluation on web browsing domain. Specifically, we focus on the following three benchmarks.
\begin{itemize}[leftmargin=2em, topsep=0pt, parsep=0pt]
    \item \textbf{WebArena} \citep{zhou2024webarena}. WebArena is a widely-adopted benchmark consisting of 812 web browsing tasks involving six domains: Shopping, Shopping Admin, Gitlab, Reddit, Map and Multisite. 
    \item \textbf{VisualWebArena} \citep{koh-etal-2024-visualwebarena}. VisualWebArena extends from WebArena and adds challenges by requiring agents to understand visual elements in the question or on the website. VisualWebArena has 910 tasks on three domains: Classifieds, Reddit and Shopping.
    \item \textbf{SCUBA} \citep{dai2025scubasalesforcecomputeruse}. SCUBA is an enterprise-centric benchmark on automating the activities of platform administrators (Admin), sales representatives (Sales), and service agents (Service) on a web-based customer relation management software. We manually excluded tasks that became invalid due to recent website updates, leaving a total of 267 tasks.
\end{itemize}

\paragraph{Evaluation practices in prior works.}
With very few exceptions \cite{yu2026polyskill,hattami2025webarena}, prior work on web agent evaluation reports pass rate over \textit{one single self-improving run} of the agent. While this is permissible for early exploration in this field and for cost considerations, recent work points out that agent performance can vary across runs \cite{yao2024taubenchbenchmarktoolagentuserinteraction,rabanser2026scienceaiagentreliability}, and hence a comprehensive re-evaluation over multiple runs is needed.
Further, when self-improvement methods are applied to web-browsing tasks \cite{wang2025agent,ouyang2025reasoningbankscalingagentselfevolving}, experiments are typically conducted under \textit{a pre-determined task order}. As our later analysis shows, this ordering induces an implicit easy-to-hard curriculum that may inadvertently favor these methods.

\paragraph{Evaluation practice in this work.} Echoing the concerns above, we broaden our evaluation along two axes: (1) To better characterize the \textit{variance}, we conduct \textit{three identical self-improving runs} for each experiment of interest. In terms of evaluation metric, we focus on \textit{run-level} statistics such as average pass rate (pass@1) per domain, along with it standard deviation across three runs, and its best-worst gap among the three runs\footnote{We are primarily interested in the variance across self-improving runs, where one \textit{run} refers to evolving over all tasks in the sequence.
\textit{Task-level} aggregated metrics such as pass@3 and pass\^{}3 are not directly applicable here; see \S\ref{app:additional-tables}.}.
(2) To investigate the influence of \textit{task order} in agent self-improving, we conduct experiments with two \textit{shuffled task orders} (\textit{i.e.}, Shuffle-1, Shuffle-2), in addition to the default order used in prior works.

\textbf{Experiment details.} We use the baseline agent harness used in \cite{prabhu2026walt} for WebArena and VisualWebArena, and the baseline agent harness used in \cite{dai2025scubasalesforcecomputeruse} for SCUBA. 
They are referred to as ``(no-memory) baseline'' in the following. 
The self-improving methods (AWM and RBank) are built on top of these baseline agents.
Unless otherwise specified, we use \texttt{GPT-5-mini} as the agent backbone model and the memory construction model. When constructing memories, we deliberately provide the ground-truth reward $r$ to the model, in contrast to prior works which uses a proxy reward $\hat{r}$ from an LLM-Judge, which we found to be noisy and further complicate our analysis\footnote{We present selected results in the main paper and defer full result tables in \S\ref{app:additional-tables}. To validate our findings beyond web-browsing tasks and \texttt{GPT-5-mini}, \S\ref{app:supplementary-experiments} introduces supplementary experiments with alternative models and benchmarks.}.

\subsection{Variance}
\label{ssec:variance}

\paragraph{Web agent evaluation is intrinsically noisy.}
In Table~\ref{tab:run-level-variance-combined}, we first report variance-related metrics for the no-memory, non-self-improving baseline in the rows marked ``Baseline''. While noise and randomness across runs are expected, it is important to quantify the magnitude of variance across runs, especially since prior work typically reports average pass@1 from a single run. In the GitLab subset of WebArena, the gap between the best and worst runs reaches a surprising 4.4\%, which in some contexts would often be considered a meaningful improvement. The standard deviation in this particular domain also reaches 2.0\%. Similarly, in VWA and SCUBA, the domain-level best-worst gap can be up to 2.4\% and 6.7\% respectively, raising concerns that single-run evaluations may be misleading and could lead to inaccurate conclusions.

\paragraph{Self-improvement methods can amplify variance.} We further compare the the rows of the baseline versus the self-improving methods in Table~\ref{tab:run-level-variance-combined}, where we see a trend that variance increases greatly when these methods are applied on top of the baseline. In 17 of the 24 cases, we see an increase in variances, and in 11 cases the relative increase exceeds 50\%.
This observation is a direct consequence of the \textit{stateful} nature of self-improving methods: memory construction is conditioned on prior task outcomes and is further influenced by stochastic LLM sampling. As a result, early randomness can compound over time, leading to substantially different memory states across runs. While this phenomenon is expected, it is concerning from an evaluation perspective, as we observe standard deviations as large as 3.9\%, and a best-worst gap also widened to 8.2\% in the Map domain and 7.8\% in the Gitlab domain of WebArena. 

\begin{table*}[t]
\centering
\caption{\textbf{Variance-related metrics across 3 runs.} Memory-based self-improving agent methods introduce increased variance in 17 out of 24 ($\approx$ 71\%) cases, among which 11 cases show a relative increase of more than 50\%.}
\label{tab:run-level-variance-combined}
\begin{threeparttable}
\centering
\scalebox{0.68}{
\begin{tabular}{l|cccccc|ccc|ccc}
\toprule
\textbf{Method} & \multicolumn{6}{c|}{\textbf{WebArena}} & \multicolumn{3}{c|}{\textbf{VisualWebArena}} & \multicolumn{3}{c}{\textbf{SCUBA}} \\
\cmidrule(lr){2-7}\cmidrule(lr){8-10}\cmidrule(lr){11-13}
 & \textbf{Shopping} & \textbf{Admin} & \textbf{GitLab} & \textbf{Map} & \textbf{Reddit} & \textbf{Multisite} & \textbf{Classifieds} & \textbf{Shopping} & \textbf{Reddit} & \textbf{Admin} & \textbf{Sales} & \textbf{Service} \\
  & (187) & (182) & (180) & (109) & (106) & (48) & (234) & (466) & (210) & (158) & (64) & (45) \\
\midrule
\rowcolor{gray!10} \multicolumn{13}{l}{Standard deviation of pass@1 across 3 runs (\%)}\\
\midrule
Baseline & 1.53 & 1.19 & 1.98 & 1.30 & 1.78 & 0.98 & 1.01 & 0.83 & 1.03 & 0.30 & 1.95 & 2.77 \\
\midrule
AWM & \cellcolor{softgreen!5}\shortstack{1.26 \\ {\scriptsize (-18\%)}} & \cellcolor{softred!50}\shortstack{2.99 \\ {\scriptsize (+152\%)}} & \cellcolor{softgreen!13}\shortstack{1.20 \\ {\scriptsize (-39\%)}} & \cellcolor{softred!17}\shortstack{1.98 \\ {\scriptsize (+53\%)}} & \cellcolor{softred!13}\shortstack{2.48 \\ {\scriptsize (+39\%)}} & \cellcolor{softgreen!0}\shortstack{0.98 \\ {\scriptsize (0\%)}} & \cellcolor{softred!25}\shortstack{1.79 \\ {\scriptsize (+78\%)}} & \cellcolor{softred!7}\shortstack{1.01 \\ {\scriptsize (+22\%)}} & \cellcolor{softred!43}\shortstack{2.38 \\ {\scriptsize (+131\%)}} & \cellcolor{softred!100}\shortstack{2.09 \\ {\scriptsize (+600\%)}} & \cellcolor{softgreen!11}\shortstack{1.28 \\ {\scriptsize (-35\%)}} & \cellcolor{softgreen!0}\shortstack{2.77 \\ {\scriptsize (0\%)}} \\
RBank & \cellcolor{softgreen!4}\shortstack{1.33 \\ {\scriptsize (-13\%)}} & \cellcolor{softred!26}\shortstack{2.12 \\ {\scriptsize (+79\%)}} & \cellcolor{softred!23}\shortstack{3.34 \\ {\scriptsize (+69\%)}} & \cellcolor{softred!66}\shortstack{3.89 \\ {\scriptsize (+200\%)}} & \cellcolor{softred!8}\shortstack{2.22 \\ {\scriptsize (+25\%)}} & \cellcolor{softred!100}\shortstack{4.28 \\ {\scriptsize (+336\%)}} & \cellcolor{softred!7}\shortstack{1.23 \\ {\scriptsize (+22\%)}} & \cellcolor{softred!30}\shortstack{1.58 \\ {\scriptsize (+91\%)}} & \cellcolor{softred!14}\shortstack{1.47 \\ {\scriptsize (+43\%)}} & \cellcolor{softred!100}\shortstack{1.58 \\ {\scriptsize (+429\%)}} & \cellcolor{softgreen!8}\shortstack{1.47 \\ {\scriptsize (-24\%)}} & \cellcolor{softred!12}\shortstack{3.78 \\ {\scriptsize (+36\%)}} \\
\midrule
\rowcolor{gray!10} \multicolumn{13}{l}{Best-worst gap of pass@1 in 3 runs (\%)}\\
\midrule
Baseline & 3.74 & 2.75 & 4.44 & 2.75 & 3.77 & 2.08 & 2.14 & 1.93 & 2.38 & 0.63 & 4.69 & 6.67 \\
\midrule
AWM & \cellcolor{softgreen!9}\shortstack{2.67 \\ {\scriptsize (-29\%)}} & \cellcolor{softred!46}\shortstack{6.59 \\ {\scriptsize (+140\%)}} & \cellcolor{softgreen!12}\shortstack{2.78 \\ {\scriptsize (-38\%)}} & \cellcolor{softred!22}\shortstack{4.59 \\ {\scriptsize (+67\%)}} & \cellcolor{softred!16}\shortstack{5.66 \\ {\scriptsize (+50\%)}} & \cellcolor{softgreen!0}\shortstack{2.08 \\ {\scriptsize (0\%)}} & \cellcolor{softred!33}\shortstack{4.27 \\ {\scriptsize (+100\%)}} & \cellcolor{softred!3}\shortstack{2.15 \\ {\scriptsize (+11\%)}} & \cellcolor{softred!46}\shortstack{5.71 \\ {\scriptsize (+140\%)}} & \cellcolor{softred!100}\shortstack{4.43 \\ {\scriptsize (+600\%)}} & \cellcolor{softgreen!11}\shortstack{3.12 \\ {\scriptsize (-33\%)}} & \cellcolor{softgreen!0}\shortstack{6.67 \\ {\scriptsize (0\%)}} \\
RBank & \cellcolor{softgreen!4}\shortstack{3.21 \\ {\scriptsize (-14\%)}} & \cellcolor{softred!26}\shortstack{4.95 \\ {\scriptsize (+80\%)}} & \cellcolor{softred!24}\shortstack{7.78 \\ {\scriptsize (+75\%)}} & \cellcolor{softred!66}\shortstack{8.26 \\ {\scriptsize (+200\%)}} & \cellcolor{softred!8}\shortstack{4.72 \\ {\scriptsize (+25\%)}} & \cellcolor{softred!100}\shortstack{10.42 \\ {\scriptsize (+400\%)}} & \cellcolor{softred!13}\shortstack{2.99 \\ {\scriptsize (+40\%)}} & \cellcolor{softred!33}\shortstack{3.86 \\ {\scriptsize (+100\%)}} & \cellcolor{softred!13}\shortstack{3.33 \\ {\scriptsize (+40\%)}} & \cellcolor{softred!100}\shortstack{3.80 \\ {\scriptsize (+500\%)}} & \cellcolor{softgreen!11}\shortstack{3.12 \\ {\scriptsize (-33\%)}} & \cellcolor{softred!11}\shortstack{8.89 \\ {\scriptsize (+33\%)}} \\
\bottomrule
\end{tabular}
}
\begin{tablenotes}
  \footnotesize\item[] Relative change vs.\ Baseline shown in parentheses. Cell color: higher (\textcolor{softred!120}{worse}) or lower (\textcolor{softgreen!120}{better}) instability.
\end{tablenotes}
\end{threeparttable}
\end{table*}

\begin{table}[t]
\centering
\begin{minipage}[t]{0.44\textwidth}
\centering
\captionof{table}{\textbf{Contextualizing our WebArena results with those in prior works.} The no-memory baseline agent in this work (based on \texttt{GPT-5-mini}) already outperforms memory-based agents in prior works.}
\label{tab:contextualizing-prior}
\vspace{0.2cm}
\scalebox{0.7}{
\begin{tabular}{l r| l r}
\toprule
\multicolumn{2}{c|}{\textbf{AWM} (Claude-3.5-Sonnet)} & \multicolumn{2}{c}{\textbf{RBank} (Gemini-2.5-Pro)} \\
\multicolumn{2}{c|}{812 Tasks} & \multicolumn{2}{c}{684 Tasks} \\
\midrule
Baseline      & 32.7 & Baseline      & 46.7 \\
+AWM          & 36.3 & +RBank        & 53.9 \\
\midrule
Our Baseline  & 54.8 & Our Baseline  & 55.3 \\
\bottomrule
\end{tabular}
}
\end{minipage}
\hfill
\begin{minipage}[t]{0.53\textwidth}
\centering
\captionof{table}{\textbf{Overall results on three benchmarks.} We report average pass@1 across 3 runs. AWM/RBank cells show score and absolute gain over the baseline (in parentheses). With stronger baselines, these methods show diminishing performance gains.}
\label{tab:aggregated-self-improve-performance-gain}
\vspace{0.1cm}
\scalebox{0.7}{
\begin{tabular}{lccc}
\toprule
\textbf{Method} & \textbf{WebArena} & \textbf{VisualWebArena} & \textbf{SCUBA} \\
\midrule
Baseline & 54.8 & 54.9 & 29.7 \\
\midrule
AWM & \cellcolor{softred!14}\shortstack{54.1 \\ {\scriptsize (-0.7)}} & \cellcolor{softred!8}\shortstack{54.5 \\ {\scriptsize (-0.4)}} & \cellcolor{softred!34}\shortstack{28.0 \\ {\scriptsize (-1.7)}} \\
RBank & \cellcolor{softgreen!30}\shortstack{56.3 \\ {\scriptsize (+1.5)}} & \cellcolor{softgreen!14}\shortstack{55.6 \\ {\scriptsize (+0.7)}} & \cellcolor{softgreen!28}\shortstack{31.1 \\ {\scriptsize (+1.4)}} \\
\bottomrule
\end{tabular}
}
\end{minipage}
\end{table}

\paragraph{Self-improvement methods have diminishing effects when a strong ``initialization'' is used.}
Since the introduction of AWM and RBank, the capabilities of language models on web browsing tasks have advanced significantly. As shown in Table~\ref{tab:contextualizing-prior}, RBank reports an average success rate of 53.9\% on a subset of 684 WebArena tasks using \texttt{Gemini-2.5-Pro}. Our baseline agent, based on the more recent GPT-5-mini model, achieves comparable performance (55.3\%) to their reported memory-enabled result (53.9\%). On VisualWebArena, our baseline agent achieves 54.9\%, which is comparable to the reported state-of-the-art result on this benchmark\footnote{The best performance on the VisualWebArena leaderboard is 54.0\% as of May 1, 2026.}.

Building on this baseline, our re-evaluation enables a reassessment of the effectiveness of these self-improvement approaches under better ``initializations''. 
As shown in Table~\ref{tab:aggregated-self-improve-performance-gain}, we find that self-improvement methods struggle to yield consistent gains.
For instance, although RBank achieves an average improvement of 1.5\% over the no-memory baseline, a two-sided Welch's $t$-test on three runs gives $p=0.23$, indicating that the gain is not statistically significant\footnote{The human performance on WebArena is 78\% \cite{zhou2024webarena}, suggesting that improvement on WebArena is still possible.}.
These results suggest that, while prior methods are effective for less capable web agents, more advanced approaches are required to construct high-quality memory and sustain the self-improvement process.

\subsection{Task Order}
Our previous results focus on broadening the evaluation through repeated, identical runs and examining the variance across those runs. These results were obtained using a default task order—from smaller task IDs to larger task IDs. In this section, we discuss the implicit assumptions underlying this design and further extend our analysis by exploring alternative task orderings.

\begin{figure}[t]
    \centering
    \includegraphics[width=1\linewidth]{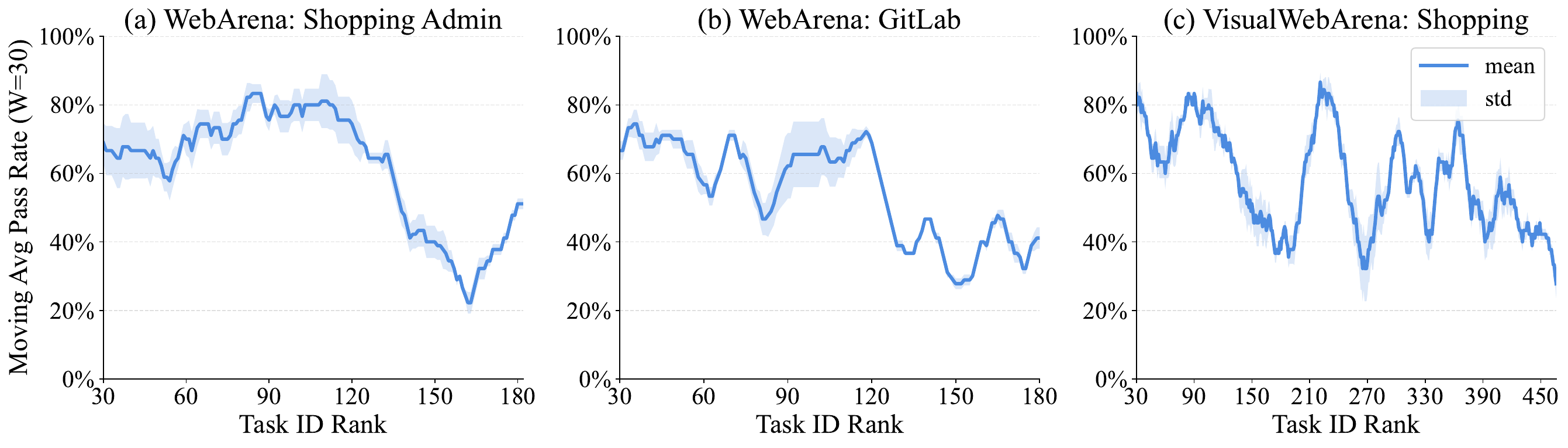}
    \caption{\textbf{Implicit curriculum in the default task order.} We visualize the moving average (window=30) of task pass rates as the no-memory baseline agent performs tasks in the default order. In (a)(b), success rates drop on tasks with larger task IDs. In (c), we observe a more complex oscillating pattern. In general, tasks of different difficulties are not distributed evenly in the default order.}
    \label{fig:implicit-task-order-examples}
\end{figure}

\paragraph{The default order exhibits an implicit easy-to-hard curriculum.}
In Figure~\ref{fig:implicit-task-order-examples}, we visualize the moving average (window = 30) of task success rates based on the no-memory baseline agent. In (a) and (b), the curves begin with high pass rates (around 75\%), which subsequently decline to below 40\% once the task ID exceeds 150. This pattern reflects an implicit easy-to-hard curriculum, where earlier tasks are more likely to fall within the agent’s capability. In (c), the sequence also begins with easier tasks, although the second half exhibits a more complex oscillatory pattern.

The default ordering may be an artifact of the benchmark construction process, whereby annotators begin with simpler tasks and progressively introduce more challenging ones.
While evaluating self-improvement approaches under this ordering is natural, real-world users may submit tasks in arbitrary orders. To stress-test self-improving systems prior to deployment, it is important to evaluate them under alternative task orderings.

\begin{figure}[t]
    \centering
    \includegraphics[width=1\linewidth]{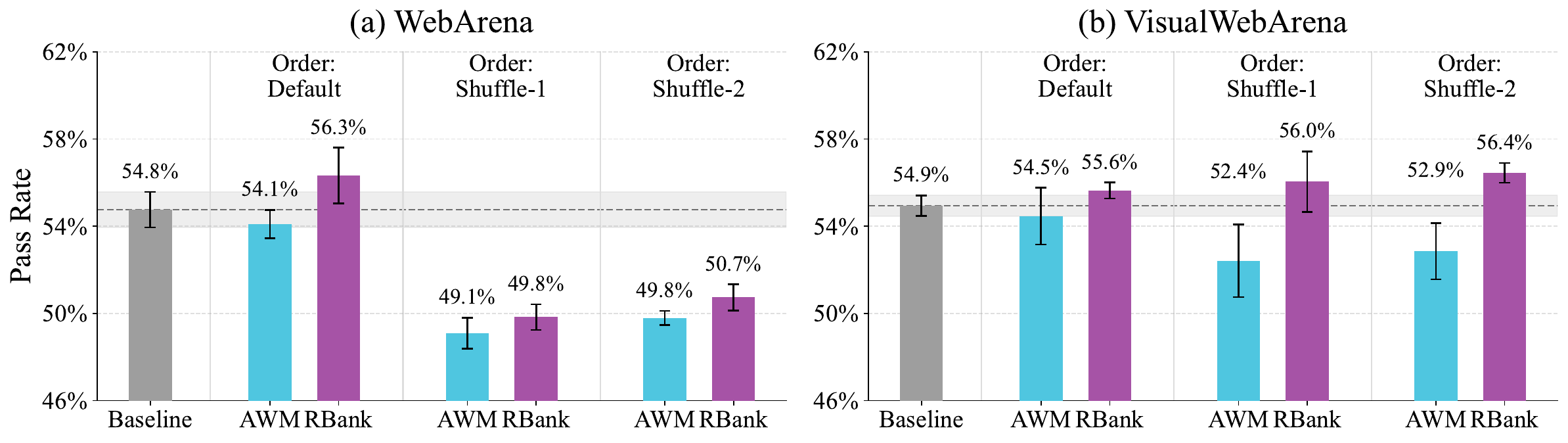}
    \caption{\textbf{Performance using different task orders.} We use the default task order and two shuffled orders. Model performance degrade significantly in 6 out of 8 cases when shuffled orders are used.}
    \label{fig:task-order-result-overview}
\end{figure}

\paragraph{Alternative task orders may lead to performance degradation.} We extend our evaluation on WebArena and VisualWebArena to two additional shuffled task orders. 
We present the result overview in Fig.~\ref{fig:task-order-result-overview}.
On WebArena, we observe a significant performance degrade when using shuffled task orders. For example, when Shuffle-1 is used, performance drops from 54.8\% to 49.1\% (AWM) and 49.8\% (RBank).
On VisualWebArena, AWM is also significantly influenced by task order, while RBank is less sensitive to this change.
While shuffled task orders create challenges for models to learn, we expect the overall success rate to be maintained at the same level as the no-memory baseline in these challenging settings; however, we see that performance often goes worse, raising concerns on the reliability of these self-improving methods in real-world settings.

\section{Indentifying Underspecification as a Potential Cause of Fragility}
\subsection{Uncovering the Failure Modes in Memory Construction}
\label{ssec:failure-modes}
To understand the amplified variance and the unexpected performance drop when shuffled orders are used, we manually inspect the agent's memory produced by the RBank method, where we identify underspecification as a potential cause of these issues. We summarize three notable findings below.

\textbf{Environment underspecification leads to recurring memories on unsupported actions.}
In our manual inspection, we identify several cases of unexpected memories that are plausible yet not applicable in the web agent's environment. For example, the agent frequently references ``API'' and recommends API-based solutions in its memory. However, the web browsing environment used in our experiments does not support such actions, and this key constraint is not provided to the memory proposal module in the RBank implementation, creating this mismatch.

\begin{figure}
    \centering
    \includegraphics[width=1\linewidth]{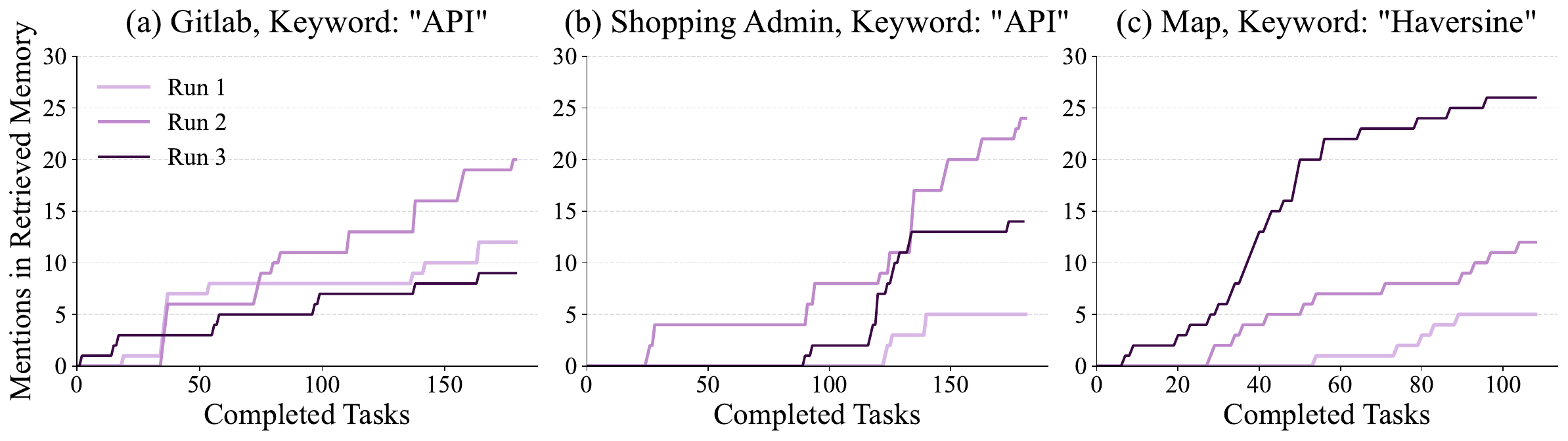}

    \caption{\textbf{Mentions of Selected Keywords in Retrieved Memory.} In (a)(b), we show that while the environment does not support API-based actions, agents increasingly save and retrieve memories about completing tasks with APIs. In (c), we show that in travel time related tasks in the map domain, the agent proposes to use Haversine Estimation instead of relying on the websites route engine, and this unintended strategy is amplified throughout the course of self-improvement.}
    \label{fig:metions-retrieved}
\end{figure}

In Fig.~\ref{fig:metions-retrieved}(a) and (b), we visualize the number of times the agent retrieves and conditions on memory items containing the keyword “API”. We observe that such mentions appear across domains, potentially introducing distraction or confusion for the agent.
Similarly, the agent frequently includes ``user confirmation'' in its memory (26 times in 3 runs of WebArena, and 22 times in 3 runs of VisualWebArena). 
While acquiring user confirmation is generally desirable, as many task queries are ambiguous, it is not supported by the evaluation environment. Consequently, such memory items may lead to situations in which the agent repeatedly issues ``wait'' actions until the time limit is reached.

\textbf{Task underspecification leads to misunderstanding, overthinking and irrelevant memories.} Task queries in WebArena are often ambiguous \cite{hattami2025webarena}, leading to unintended behaviors from the agent. 
For example, in WebArena task 118, the agent is tasked with ``I have jaw bruxism problem, show me something that could alleviate the problem.'' The intended goal is for the agent to locate a relevant product (\textit{i.e.}, a mouth guard) on the shopping website; however, the agent instead adopts a more literal interpretation and responds with ``consult your dentist or doctor for personalized advice.''
Further, when the agent is prompted to reflect on its failure on this task, it produces irrelevant memories such as ``gather targeted patient/context details before giving medical guidance.'' We also observe cases in which benchmark evaluator bugs lead to false negatives during evaluation, which in turn cause the agent to overthink and generate spurious memories. See Table~\ref{tab:ambiguous-task-query-example}-\ref{tab:evaluator-bug-example} for concrete examples.

\textbf{Memories are ``contagious'': the case of using Haversine Formula in map tasks.} 
Surprisingly, in the map domain of WebArena, we observe references to the ``Haversine Formula'', a method for computing the shortest distance between two points on a sphere. 
Upon further inspection, we find that when the agent is tasked with obtaining the distance between two locations, the map website does not always load as expected or respond in time.
As a result, the agent adopts a fallback strategy of estimating distances using the Haversine Formula. This approximation occasionally yields correct answers (see Table~\ref{tab:duquesne-one-hour-eval} for a detailed example), which leads the agent to reinforce these unintended strategies into its memory.
As shown in Fig.~\ref{fig:metions-retrieved}, the introduction of this strategy appears to be stochastic, and the earlier ``Haversine'' is added to memory, the more frequently it is retrieved and used in later tasks. This effect may contribute to increased variance across runs.

\label{sec:combacting-fragility}

\definecolor{gitgreenbg}{RGB}{230, 255, 237}   
\definecolor{gitgreenfg}{RGB}{31, 136, 61}     
\begin{table}
\small
    \caption{\textbf{Three types of additional information used in \S\ref{ssec:additional-info}.} We hypothesize such information can reduce underspecification when constructing memories. Examples are truncated due to space limit.}
    \label{tab:improved-specification}
    \centering
    \begin{tabularx}{\linewidth}{X}
    \toprule
    \rowcolor{gray!10}Rubrics and Scores (+Rub): \\\midrule
    Evaluator: string\_match | Score: 0.0 (FAIL)\\
    Prediction: "Top-2 best-selling products in 2022: 1) Quest Lumaflex™ Band, 5 units; 2) Cruise Dual Analog Watch, 4 units."\\
    must\_include: "Quest Lumaflex™ Band" → PASS\\
    must\_include: "Sprite Stasis Ball 65 cm" → FAIL\\
    \midrule
    \rowcolor{gray!10}Environment Feedback (+Env): \\\midrule
    \textcolor{gray!75}{Actions: click(31), input\_text(41, ``05/01/2022''), 
    input\_text(44, ``12/31/2022''), click(59) {\tiny\# Preceding Context}}\\
    Action error: Error executing action input\_text: Failed to input text into index 41\\\midrule
    \rowcolor{gray!10} Prompt Modification (+PMod): \\\midrule
    - \textbf{INCLUDE}: Procedural knowledge, reusable workflows, site navigation structure, UI interaction tips. \\
    - \textbf{STRICTLY AVOID}: API or programmatic solutions, visiting external websites, requesting human confirmation, or task-specific details tied to a particular site or query. \\
    \bottomrule
    \end{tabularx}
\end{table}

\subsection{Addressing Underspecification with Additional Information}
\label{ssec:additional-info}
In the original implementation of RBank, the memory construction module is minimal and generic; \textit{i.e.}, ``You need to extract and summarize useful insights in the format of memory items based on the agent’s successful trajectory'' (see Appendix~\ref{app:prompts} for the full prompt). Based on our findings in \S\ref{ssec:failure-modes}, such generic descriptions may be insufficient, particularly when the baseline agent is already strong and more specific feedback or memory is required to achieve meaningful improvements.

In the following, we address this limitation by incorporating additional information during memory construction. We describe three types of such information below and provide examples in Table~\ref{tab:improved-specification}.

\begin{itemize}[leftmargin=2em, topsep=0pt, parsep=0pt]

\item \textbf{Rubrics and Scores (+Rub).} Across all considered benchmarks, tasks are evaluated using function-based rubrics. For example, WebArena adopts three types of evaluation functions (\texttt{must\_include}, \texttt{exact\_match}, and \texttt{fuzzy\_match}), which can be applied to the agent’s returned answer or final state. In the jaw bruxism example discussed earlier (Table~\ref{tab:ambiguous-task-query-example}), we demonstrate that an ambiguous task query can lead to multiple interpretations and outcomes. Providing rubrics can help clarify the intended task and improve the relevance of the resulting memory. Note that these rubrics are provided only during post-completion memory construction; \textit{i.e.}, the agent remains unaware of them while performing the task.

\item \textbf{Environment Feedback (+Env).} We incorporate web environment feedback, such as whether click or selection actions are successful, into the memory construction process. In practice, we observe cases where the agent issues a sequence of actions (\textit{e.g.}, click item 31, then fill textbox 41). However, after clicking item 31, textbox 41 may no longer be present on the page, resulting in an action error. Such information is not included in the original memory generation process, which can lead the agent to repeat similar UI interaction mistakes. Incorporating this feedback provides a useful signal for improving performance in future tasks.

\item \textbf{Prompt Modification (+PMod).}
Our previous analysis shows that the agent may ``learn'' unintended strategies, such as proposing API-based actions or requesting confirmation from human users. While these behaviors are plausible, they are not supported in the agent environment. To address this, we introduce a prompt edit that explicitly discourages such strategies. Additionally, we encourage the inclusion of procedural knowledge and site navigation structures, thereby guiding the model to generate memories that are more relevant to web-browsing tasks.

\item \textbf{+All.} Finally, we experiment with including all three modifications mentioned above. 

\end{itemize}

\begin{figure}[t]
    \centering
    \includegraphics[width=0.9\linewidth]{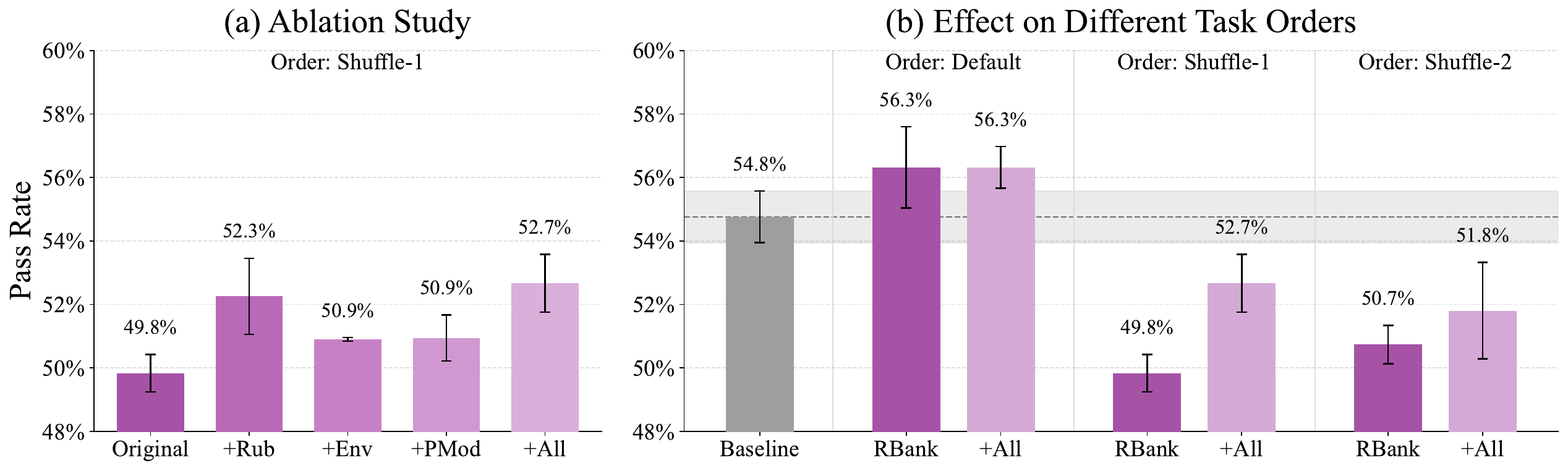}
    \caption{\textbf{Addressing Underspecification with Additional Information.} (a) We apply the three types of information individually and altogether. (b) We experiment with +All using three task orders.}
    \label{fig:additional-info-for-memory-construction}
    \vspace{-0.3cm}
\end{figure}

\textbf{Results and Discussion.} 
We begin by applying these modifications to the RBank experiments under the Shuffle-1 task order. The results are presented in Fig.~\ref{fig:additional-info-for-memory-construction}(a), where we observe that each type of additional information yields modest performance gains. When combined, these modifications result in a 2.9\% improvement over the original RBank method (49.8\% $\rightarrow$ 52.7\%).
We further evaluate the +All setting under the Default and Shuffle-2 orders, with results shown in Fig.~\ref{fig:additional-info-for-memory-construction}(b). We observe an improvement of 1.1\% on Shuffle-2, while performance under the Default order is maintained.

Overall, these findings suggest that incorporating additional information during memory construction is beneficial and non-disruptive; thus, when such information is available, its incorporation during memory construction is recommended. 
However, even with these enhancements, the agent still underperforms the no-memory baseline under Shuffle-1 and Shuffle-2. 
Indeed, the three types of additional information represent our initial attempt to address this issue and may not fully capture all sources of underspecification. Other factors contributing to robustness under varying task orders may also exist and remain unidentified. We leave further investigation of this issue to future work.

\section{Related Works}

\textbf{Agent Continual Learning and Self-Improvement.} 
This work focuses on an inference-only setting for agent self-improvement, motivated by its practical utility, accessible interface, and as a way to explore the limits of agents in forward-pass only settings. Beyond web-browsing tasks, related self-improvement methods have been studied for math, reasoning, and coding agents \citep{hu2025automated,zhang2026darwin,wang2026huxleygodel}, as well as long-term personalization \citep{maharana-etal-2024-evaluating,shaikh2026learningactionpredictorshumancomputer}.
More broadly, continual learning and self-improvement (or ``self-play'') have long been studied prior to the inference-only paradigm, including in supervised fine-tuning \citep{zelikman2022star, chen2024selfplay} and reinforcement learning settings \citep{alphago}.

\textbf{Agent Skill and Memory.}
One promising direction for agent continual learning is the construction of ``memories'', including both parametric and non-parametric forms. This work focuses on non-parametric, textual memory, which has gained significant traction recently \cite{AMEM, personamem, Memorybank, MemGPT, mem0}. Relatedly, workflow-related memories are sometimes referred to as ``skills'' and has been extensively studied~\cite{zhang2025skills,li2026skillsbenchbenchmarkingagentskills,tian2026skillscoachselfevolvingskilloptimizer,jiang2026xskillcontinuallearningexperience}.
For parametric memory, representative approaches include sparse memory fine-tuning \cite{sparsefinetuning}, which updates trainable memory slots based on their relative activation on unseen tasks compared to a background corpus. Additionally, memory mechanisms have been studied from a model architecture perspective, with notable works including Hope \cite{NestedLearning} and Titans \cite{TITANS}.

\textbf{Agent Evaluation and Reliability.}
The research community has developed diverse benchmarks for evaluating agent capabilities, in domains such as tool use \cite{schick2023toolformerlanguagemodelsteach,wang2024executable}, computer use \cite{OSWorld}, web browsing \cite{zhou2024webarena,koh-etal-2024-visualwebarena,deng2023mind2web}. 
As agent capabilities continue to grow, newer benchmarks target tasks with real-world economical values \cite{patwardhan2025gdpvalevaluatingaimodel}, including coding \cite{jimenez2024swebench,deng2025swebenchproaiagents}, customer service \cite{yao2024taubenchbenchmarktoolagentuserinteraction,barres2025tau2benchevaluatingconversationalagents}, customer relationship management \cite{huang-etal-2025-crmarena,huang2026crmarenapro}, and enterprise operations \citep{malay2026enterpriseopsgym}.
Along these contributions, there is also growing concern regarding the evaluation sciences of agents \cite{rabanser2026scienceaiagentreliability,grace2026demystifying,zhu2025establishingbestpracticesbuilding,kapoor2025holisticagentleaderboardmissing}. 
In particular, \citet{rabanser2026scienceaiagentreliability} advocate treating reliability as a primary evaluation axis alongside capability, a perspective we strongly agree with. We extend this viewpoint from single-session agents to multi-session self-improving agents, where reliability becomes even more critical and harder to measure.

\vspace{-0.2cm}
\section{Conclusion}
\label{sec:conclusion}
In this work, we conducted a comprehensive re-evaluation of two established memory-based methods for agent self-improvement, broadening the scope of evaluation along two axes (multiple runs and shuffled task orders) while using a stronger combination of base model and agent harness as the baseline (\textit{i.e.}, the starting point for self-improvement). 
Overall, our results reveal the fragility of these self-improving systems under more challenging settings: the results of self-improving experiments can exhibit significant variance, meaning that evaluation based on a single run can hinder accurate measurement of progress. 
Furthermore, agents do not self-improve, but instead degrade, in challenging settings where the baseline is already strong or when tasks arrive in a random order.

Given these surprising findings, we further manually examine the agents' memory and identify environment and task underspecification as a key limitation in existing approaches, supporting this with detailed case studies and analyses. We address this limitation by incorporating additional information that provides better specification to the memory generation module. By doing so, the performance degradation observed in prior experiments is reduced by 31\%, demonstrating promising progress. However, the remaining 69\% performance gap suggests that there are still fundamental issues to be uncovered and resolved to fully overcome this fragility.

Beyond these empirical observations, our findings offer practical recommendations for researchers and practitioners working with self-improving agentic systems.

\textbf{Recommendations for Evaluation.}
Our work advocates for more rigorous evaluation protocols that account for the stability issues discussed in this study. Specifically, results from a single run should be interpreted with caution; instead, reporting outcomes across multiple self-improving runs and randomized task orders should be highly encouraged. Furthermore, we recommend first piloting new methods on benchmarks with well-specified tasks, followed by stress-testing them on underspecified tasks to estimate worst-case performance prior to deployment.

\textbf{Recommendations for System Development.}
Our analysis reveals that without proper validation mechanisms, agent memories are merely unverified hypotheses rather than true lessons learned. They can be erroneous and, more importantly, cascade negatively into subsequent tasks over time. Future work should investigate memory validation mechanisms that filter out problematic memories to ensure long-term robustness.
Additionally, as demonstrated in \S\ref{ssec:failure-modes}, task and environment specification is a nuanced challenge. Because underspecification cannot be fully foreseen, and real-world deployments unfold as a single, irreversible run, refining specifications through an iterative process represents one viable path forward. Consequently, our findings motivate the development of interfaces that enable effective human oversight and timely intervention before agents fail unexpectedly.






\bibliographystyle{unsrtnat}
\bibliography{custom} 


\appendix
\section{Limitations}
\label{app:limitations}

Our evaluation in the main paper focuses on a specific class of memory-based self-improving agents applied to web-browsing tasks with a single model. 
To keep the experimental design tractable, we did not extend the core analysis to other types of self-improvement methods, other domains, or a wider range of models, and given the rapid pace of development in this field, we could not evaluate the most recently introduced memory management techniques. 
To partially address these limitations, we conducted additional experiments in the appendix (\S\ref{app:supplementary-experiments}: two additional models, a filesystem-based self-improving method, an enterprise tool-use benchmark, and a larger number of self-improving runs). These experiments broadly corroborate our main findings. 
Our evaluation is far from exhaustive. However, since the methods we evaluate are minimalist, highly scalable, and increasingly adopted, we believe the issues identified  are broadly representative.

Secondly, our manual inspection of agent memories was qualitative and covered only a subset of the memories produced. Given the sheer volume of memory entries generated in our experiments, an exhaustive review was infeasible. While our analysis highlights several critical failure modes, other undocumented patterns likely exist. Future work would benefit from analyzing agent memories at scale and developing interfaces for human review of a large volume of agent memories.

Lastly, underspecification only explains part of the performance degradation, and the reasons behind the remaining performance gap are still unknown.
It likely arises from multiple interacting factors: the quality of the underlying tasks, the design of the self-improvement and memory mechanisms, and the characteristics of the base LLM. Systematically attributing the degradation to these factors would require the scalable memory analysis pipeline described above, benchmark re-annotation efforts, and extensive experiments to isolate the effect of each factor. We leave this for future work.

\section{Extended Related Works}

\textbf{Monitorability.} Our work is also broadly related to literature on chain-of-thought and agent monitorability \cite{baker2025monitoringreasoningmodelsmisbehavior,guan2025monitoringmonitorability,meng2025docent}. For instance, \citet{baker2025monitoringreasoningmodelsmisbehavior} observe instances of reward hacking in coding agents via chain-of-thought monitoring. This aligns closely with our findings, as we observe similar cases of ``reward hacking'' (\textit{e.g.}, the Haversine Formula example in Table~\ref{tab:duquesne-one-hour-eval}) occurring during the agent's learning process. 
Because memory-based continual learning provides an accessible textual interface to inspect an agent's ``learning,'' it introduces opportunities to explore scalable monitoring methods, as well as interfaces to intervene in the process.

\textbf{Reliability and Stability Concerns of Emerging Learning Paradigms.}
Reliability has been a recurring challenge across evolving learning paradigms in AI. Prior works have extensively investigated the stability of deep reinforcement learning~\citep{henderson2019deepreinforcementlearningmatters}, gradient-based meta-learning~\citep{finn2017modelagnosticmetalearningfastadaptation,antoniou2018how}, language model fine-tuning~\citep{zhang2021revisiting,dodge2020finetuningpretrainedlanguagemodels}, prompting~\citep{sclar2024quantifying,multi-prompt-eval}, in-context learning~\citep{zhao2021calibrateuseimprovingfewshot,lu-etal-2022-fantastically,voronov-etal-2024-mind}, and self-correction~\citep{huang2024large}.
Self-improving agents represent a nascent paradigm enabled by increasingly capable language models. However, unlike earlier paradigms confined to passive prediction or generation, modern agents are increasingly integrated into real-world systems where they are granted the autonomy to execute irreversible actions. Consequently, deploying them without systematically addressing underlying fragility risks not only compounding computational errors, but tangible real-world harm, making reliability an urgent prerequisite for safe adoption.

\section{Additional Experiment Details}
\label{app:additional-experiment-details}
\paragraph{Costs.} When using \texttt{GPT-5-mini} as the backbone model, a single run over all 812 WebArena tasks costs around \$25, and a single run over all 267 SCUBA tasks costs \$29. These estimates are based on the no-memory baselines. For memory-based methods, additional costs arise from (a) generating memories via extra LM calls and (b) including memories in the context. However, we expect the additional cost from both components to be minimal. All experiment code are hosted and run from a standard 64-CPU server.

\paragraph{LLM configurations.} When using \texttt{GPT-5-mini}, we use the default (medium) reasoning efforts and the default temperature (1.0) during generation. 

\paragraph{Dataset and Code License.} The WebArena benchmark \cite{zhou2024webarena} is released under Apache-2.0 license. The VisualWebArena benchmark \cite{koh-etal-2024-visualwebarena} is released under MIT license. The SCUBA benchmark \cite{dai2025scubasalesforcecomputeruse} is released under Apache-2.0 license. Our baseline agents were revised from WALT \cite{prabhu2026walt} and SCUBA \cite{dai2025scubasalesforcecomputeruse}, released under MIT license and Apache-2.0 license respectively. In \S\ref{app:supplementary-experiments}, we conducted additional experiments using the EnterpriseOps-Gym benchmark \cite{malay2026enterpriseopsgym} and the Letta Code framework \cite{MemGPT,lin2025sleeptimecomputeinferencescaling}. Both are both released under Apache-2.0 license.

\paragraph{LLM Usage.} We used coding agents to support our experimentation. Our code is largely based on existing implementation from prior works, therefore these coding agents do not contribute original, new ideas. We also used LLMs for paper editing and proofreading.


\newpage

\section{Result Tables}
Due to space constraints, the main paper reports only aggregated metrics (\textit{e.g.}, pass@1 averaged over domains; Table~\ref{tab:aggregated-self-improve-performance-gain}). In this section, we present results under alternative metrics (pass@k and pass\^{}k) for the no-memory baseline, as well as a per-domain breakdown for all compared methods.

\label{app:additional-tables}
\subsection{Pass@3 and Pass\^{}3 Metrics for the No-Memory Baseline}

For the no-memory baseline, each task in the benchmark is performed \textit{independently}, so pass@k and pass\^{}k \citep{yao2024taubenchbenchmarktoolagentuserinteraction} are natural metrics for capturing \textit{task-level} variance across repeated executions of the same task. Note that this differs from the \textit{run-level} variance discussed in the main paper, which measures variance across multiple self-improving runs.

We report both metrics in Table~\ref{tab:wa-baseline-pass-metrics}-\ref{tab:vwa-scuba-baseline-pass-metrics}. Consistent with prior findings, agents are often quite inconsistent across runs: the gap between pass@3 and pass\^{}3 frequently exceeds 20\%.

\begin{table}[h]
\centering
\caption{Evaluation of the no-memory baseline agent on WebArena.}
\label{tab:wa-baseline-pass-metrics}
\scalebox{0.9}{
\begin{tabular}{l|cccccc|c}
\toprule
\textbf{Metric} & \textbf{Shopping} & \textbf{Admin} & \textbf{GitLab} & \textbf{Map} & \textbf{Reddit} & \textbf{Multisite} & \textbf{All} \\
 & (187) & (182) & (180) & (109) & (106) & (48) & (812) \\
\midrule
Avg.\ Pass & 45.8 & 62.6 & 53.9 & 53.2 & 69.5 & 34.0 & 54.8 \\
Pass@3 & 55.1 & 72.0 & 63.3 & 67.0 & 82.1 & 47.9 & 65.4 \\
Pass\^{}3 & 34.8 & 52.2 & 45.0 & 40.4 & 53.8 & 20.8 & 43.3 \\
\bottomrule
\end{tabular}
}
\end{table}

\begin{table}[h]
\centering
\caption{Evaluation of the no-memory baseline agent on VisualWebArena and SCUBA.}
\label{tab:vwa-scuba-baseline-pass-metrics}
\scalebox{0.9}{
\begin{tabular}{l|ccc|c|ccc|c}
\toprule
\textbf{Metric} & \multicolumn{4}{c|}{\textbf{VisualWebArena}} & \multicolumn{4}{c}{\textbf{SCUBA}} \\
\cmidrule(lr){2-5}\cmidrule(lr){6-9}
 & \textbf{Classifieds} & \textbf{Shopping} & \textbf{Reddit} & \textbf{All} & \textbf{Admin} & \textbf{Sales} & \textbf{Service} & \textbf{All} \\
 & (234) & (466) & (210) & (910) & (158) & (64) & (45) & (267) \\
\midrule
Avg.\ Milestone & -&-&-&-& 46.7 & 61.8 & 42.3 & 49.6 \\
Avg.\ Pass & 60.1 & 57.6 & 43.3 & 54.9 & 26.2 & 39.6 & 28.1 & 29.7 \\
Pass@3 & 73.9 & 69.1 & 51.4 & 66.3 & 39.2 & 57.8 & 40.0 & 43.8 \\
Pass\^{}3 & 44.9 & 45.1 & 33.8 & 42.4 & 13.9 & 20.3 & 13.3 & 15.4 \\
\bottomrule
\end{tabular}
}
\end{table}

\subsection{Per-domain Results}

For the memory-based methods (AWM and RBank), we conduct multiple self-improving runs, and task execution within a given run \textit{depends} on the memory accumulated in that run. Consequently, pass@3 and pass\^{}3 are not directly applicable. Instead, we track the pass rate (pass@1) over the sequence of tasks within each self-improving run and report the mean, standard deviation, and best–worst gap across the three runs. For consistency, we report the same metrics for the no-memory baseline, even though it does not self-improve.

Results are presented in Tables~\ref{tab:wa-per-domain}--\ref{tab:scuba-release-per-domain-binary}, which also serve as the per-domain breakdown of Table~\ref{tab:aggregated-self-improve-performance-gain} in the main paper.

We note that one could additionally evaluate each task three times within each of the three self-improving runs to measure pass@3/pass\^{}3. However, this would triple the cost of our evaluation protocol and cost nine times as much as the standard single-run evaluation, which is prohibitive in this work.

\begin{table*}[h]
\centering
\small

\caption{\textbf{Per-domain pass rate (\%) on WebArena.} Each cell reports mean/std/best--worst gap, based on three runs of memory construction. The right-most column (``All'') reports the overall pass rate.}
\label{tab:wa-per-domain}
\setlength{\tabcolsep}{4pt}
\scalebox{0.9}{%
\begin{tabular}{lccccccc}
\toprule
\textbf{Method} & \textbf{Shopping} & \textbf{Admin} & \textbf{GitLab} & \textbf{Map} & \textbf{Reddit} & \textbf{Multisite} & \textbf{All} \\
 & (187) & (182) & (180) & (109) & (106) & (48) & (812) \\
\midrule
No Memory & 45.8/1.53/3.74 & 62.6/1.19/2.75 & 53.9/1.98/4.44 & 53.2/1.30/2.75 & 69.5/1.78/3.77 & 34.0/0.98/2.08 & 54.8 \\
\midrule
\rowcolor{gray!10}\multicolumn{8}{l}{Task Order: Ordinal} \\\midrule
AWM & 48.3/1.26/2.67 & 59.7/2.99/6.59 & 51.7/1.20/2.78 & 50.5/1.98/4.59 & 72.3/2.48/5.66 & 32.6/0.98/2.08 & 54.1 \\
RBank & 52.8/1.33/3.21 & 60.6/2.12/4.95 & 59.3/3.34/7.78 & 48.6/3.89/8.26 & 71.1/2.22/4.72 & 27.8/4.28/10.42 & 56.3 \\
\midrule
\rowcolor{gray!10}\multicolumn{8}{l}{Task Order: Shuffle-1} \\\midrule
AWM & 43.3/1.51/3.21 & 54.8/2.70/6.59 & 46.7/1.36/3.33 & 44.0/2.70/6.42 & 71.1/2.22/4.72 & 22.2/3.93/8.33 & 49.1 \\
RBank & 42.4/1.10/2.67 & 56.0/1.55/3.30 & 45.7/0.94/2.22 & 46.8/2.25/5.50 & 72.6/2.04/4.72 & 27.1/0.00/0.00 & 49.8 \\
\midrule
\rowcolor{gray!10}\multicolumn{8}{l}{Task Order: Shuffle-2} \\\midrule
AWM & 43.5/1.40/3.21 & 55.1/1.13/2.75 & 44.6/2.92/6.67 & 48.6/0.75/1.83 & 69.5/1.60/3.77 & 32.6/3.93/8.33 & 49.8 \\
RBank & 44.4/1.16/2.67 & 56.8/2.63/6.04 & 46.3/2.77/6.67 & 51.7/1.14/2.75 & 68.2/1.60/3.77 & 28.5/0.98/2.08 & 50.7 \\
\bottomrule
\end{tabular}}
\end{table*}

\begin{table*}[t]
\centering
\small
\caption{\textbf{Per-domain pass rate (\%) on VisualWebArena.} Each cell reports mean/std/best--worst gap, based on three runs of memory construction. The right-most column (``All'') reports the overall pass rate.}
\label{tab:vwa-per-domain}
\setlength{\tabcolsep}{4pt}
\scalebox{0.9}{%
\begin{tabular}{lcccc}
\toprule
\textbf{Method} & \textbf{Classifieds} & \textbf{Shopping} & \textbf{Reddit} & \textbf{All} \\
 & (234) & (466) & (210) & (910) \\
\midrule
No Memory & 60.1/1.01/2.14 & 57.6/0.83/1.93 & 43.3/1.03/2.38 & 54.9 \\
\midrule
\rowcolor{gray!10}\multicolumn{5}{l}{Task Order: Ordinal} \\\midrule
AWM & 61.3/1.79/4.27 & 55.3/1.01/2.15 & 45.1/2.38/5.71 & 54.5 \\
ReasoningBank & 59.5/1.23/2.99 & 58.3/1.58/3.86 & 45.4/1.47/3.33 & 55.6 \\
\midrule
\rowcolor{gray!10}\multicolumn{5}{l}{Task Order: Shuffle 1} \\\midrule
AWM & 57.3/0.92/2.14 & 53.2/3.04/6.65 & 45.2/0.67/1.43 & 52.4 \\
ReasoningBank & 60.3/2.09/5.13 & 58.9/2.33/5.15 & 45.1/0.22/0.48 & 56.0 \\
\midrule
\rowcolor{gray!10}\multicolumn{5}{l}{Task Order: Shuffle 2} \\\midrule
AWM & 58.4/2.13/5.13 & 54.9/0.63/1.50 & 42.1/2.76/6.67 & 52.9 \\
ReasoningBank & 62.5/0.53/1.28 & 58.7/0.27/0.64 & 44.8/2.43/5.71 & 56.4 \\
\bottomrule
\end{tabular}}
\end{table*}

\begin{table}[h]
\centering
\small
\caption{\textbf{Per-domain milestone score (\%) on SCUBA.} Each cell reports mean/std/best--worst gap, based on three runs of memory construction. The right-most column (``All'') reports the overall pass rate.}
\label{tab:scuba-per-domain}
\setlength{\tabcolsep}{4pt}
\scalebox{0.9}{%
\begin{tabular}{lcccc}
\toprule
\textbf{Method} & \textbf{Admin} & \textbf{Sales} & \textbf{Service} & \textbf{All} \\
 & (158) & (64) & (45) & (267) \\
\midrule
Baseline & 46.7/0.25/0.59 & 61.8/1.57/3.36 & 42.3/1.40/3.02 & 49.6 \\
Synapse~\citep{zheng2024synapse} & 51.3/1.39/3.40 & 64.9/2.64/5.94 & 48.6/5.24/12.51 & 54.1 \\
AWM & 45.8/1.40/3.25 & 60.8/2.98/7.19 & 49.6/4.15/9.13 & 50.1 \\
ReasoningBank & 50.7/0.19/0.47 & 62.4/2.70/6.59 & 36.3/2.85/6.11 & 51.1 \\
\bottomrule
\end{tabular}}
\end{table}

\begin{table}[t]
\centering
\small

\caption{\textbf{Per-domain pass rate (\%) on SCUBA (binary success: milestone score $=1.0$).} Each cell reports mean/std/best--worst gap, based on three runs of memory construction. The right-most column (``All'') reports the overall pass rate.}
\label{tab:scuba-release-per-domain-binary}
\setlength{\tabcolsep}{4pt}
\scalebox{0.9}{%
\begin{tabular}{lcccc}
\toprule
\textbf{Method} & \textbf{Admin} & \textbf{Sales} & \textbf{Service} & \textbf{All} \\
 & (158) & (64) & (45) & (267) \\
\midrule
Baseline & 26.2/0.30/0.63 & 39.6/1.95/4.69 & 28.1/2.77/6.67 & 29.7 \\
Synapse~\citep{zheng2024synapse} & 30.4/1.37/3.16 & 43.2/2.66/6.25 & 28.9/1.81/4.44 & 33.2 \\
AWM & 22.4/2.09/4.43 & 37.5/1.28/3.12 & 34.1/2.77/6.67 & 28.0 \\
ReasoningBank & 30.2/1.58/3.80 & 40.1/1.47/3.12 & 21.5/3.78/8.89 & 31.1 \\
\bottomrule
\end{tabular}}
\end{table}
\clearpage

\section{Supplementary Experiments}

\begin{table}[h]
\centering
\small
\caption{\textbf{WebArena Results, Three Additional Runs using \texttt{GPT-5-mini}.} Each cell reports avg/std/best-worst gap across $n$ runs, and the right-most column (``All'') reports the overall pass rate. }
\label{tab:additional-runs}
\setlength{\tabcolsep}{4pt}
\scalebox{0.9}{
\begin{tabular}{lccccccc}
\toprule
\textbf{Method} & \textbf{Shopping} & \textbf{Admin} & \textbf{GitLab} & \textbf{Map} & \textbf{Reddit} & \textbf{Multisite} & \textbf{All} \\
 & (187) & (182) & (180) & (109) & (106) & (48) & (812) \\\midrule

\rowcolor{gray!10}\multicolumn{8}{l}{\textit{$n = 3$ runs}} \\
\midrule
No Memory      & 45.8/1.5/3.7 & 62.6/1.2/2.8 & 53.9/2.0/4.4 & 53.2/1.3/2.8 & 69.5/1.8/3.8 & 34.0/1.0/2.1 & 54.8 \\
RBank (Ordinal) & 52.8/1.3/3.2 & 60.6/2.1/5.0 & 59.3/3.3/7.8 & 48.6/3.9/8.3 & 71.1/2.2/4.7 & 27.8/4.3/10.4 & 56.3 \\
\midrule
\rowcolor{gray!10}\multicolumn{8}{l}{\textit{$n = 6$ runs}} \\
\midrule
No Memory      & 45.8/1.7/4.3 & 61.7/1.8/5.5 & 53.7/1.5/4.4 & 52.0/2.2/6.4 & 68.1/2.7/7.6 & 34.0/1.0/2.1 & 54.2 \\
RBank (Ordinal) & 51.1/2.5/8.0 & 60.8/1.9/5.0 & 58.3/2.8/8.9 & 47.3/3.1/9.2 & 71.4/2.3/6.6 & 28.1/3.9/10.4 & 55.6 \\
\bottomrule
\end{tabular}
}
\end{table}

\label{app:supplementary-experiments}
\paragraph{Additional Runs (n=3 $\rightarrow$ n=6).} In the main paper, we evaluated each memory-based method with three self-improving runs. To characterize the benefits of further increasing n, we conduct three additional runs on a subset of experiments (WebArena; no-memory baseline and ReasoningBank; ordinal order), for n=6 in total. In Table~\ref{tab:additional-runs} we report the original (n=3) and extended (n=6) results. 

We find that the spread widens from n=3 to n=6: best-worst gaps grow in most domains (\textit{e.g.}, shopping: 3.21$\rightarrow$8.02 for ReasoningBank), indicating that n=3 may understate the true variability, and larger n better characterizes the spread and worst-case behavior.
The variance-amplification finding on \texttt{GPT-5-mini} holds at n=6;  ReasoningBank shows a larger standard deviation than the no-memory baseline in 5 of 6 domains.

Finally, we perform Welch's $t$-tests comparing the no-memory baseline and ReasoningBank using the n=6 runs. On the overall success rate, this yields $p=0.07$, which is lower than the $p=0.23$ obtained with only three runs but still not significant at $\alpha=0.05$; ReasoningBank's overall gain in this setting therefore cannot be distinguished from run-to-run noise.
Per-domain results are mixed, with statistically significant differences in both directions for some domains.

\begin{table}[h]
\centering
\small
\caption{\textbf{WebArena Results using Additional Models.} Each cell reports avg/std/best-worst gap across three runs, and the right-most column (``All'') reports the overall pass rate. }
\label{tab:additional-models}
\setlength{\tabcolsep}{4pt}
\scalebox{0.9}{
\begin{tabular}{lccccccc}
\toprule
\textbf{Method} & \textbf{Shopping} & \textbf{Admin} & \textbf{GitLab} & \textbf{Map} & \textbf{Reddit} & \textbf{Multisite} & \textbf{All} \\
 & (187) & (182) & (180) & (109) & (106) & (48) & (812) \\
\midrule
\rowcolor{gray!10}\multicolumn{8}{l}{\texttt{GPT-OSS-120B}} \\
\midrule
No Memory      & 35.5/1.1/2.7 & 46.7/0.4/0.8 & 45.4/0.9/2.2 & 33.0/1.5/3.7 & 45.9/2.5/5.7  & 24.3/3.9/8.3 & 40.6 \\
RBank (Ordinal)  & 38.0/0.9/2.1 & 49.0/1.7/4.1 & 42.4/0.3/0.6 & 30.0/2.2/4.6 & 48.1/5.4/13.2 & 18.8/1.7/4.2 & 40.5 \\
RBank (Shuffle-1) & 34.3/1.5/3.2 & 40.4/2.6/5.8 & 34.5/1.7/3.9 & 33.9/2.2/5.5 & 49.4/1.2/2.8  & 22.9/3.4/8.3 & 37.0 \\
\midrule
\rowcolor{gray!10}\multicolumn{8}{l}{\texttt{Gemini-2.5-Pro}} \\
\midrule
no-memory        & 40.3/1.1/2.7 & 59.2/2.4/5.2 & 48.7/2.1/5.0 & 33.9/1.3/2.8 & 70.4/2.4/5.7 & 33.3/3.4/8.3 & 49.0 \\
RBank (ordinal)  & 42.8/1.2/2.7 & 59.3/1.6/3.8 & 51.1/4.3/9.4 & 37.0/1.6/3.7 & 75.2/1.6/3.8 & 30.6/2.6/6.3 & 51.0 \\
RBank (shuffle1) & 38.3/2.8/6.4 & 57.2/0.6/1.4 & 44.2/1.6/3.9 & 37.0/3.0/7.3 & 73.9/1.9/4.7 & 32.6/1.0/2.1 & 48.0 \\
\bottomrule
\end{tabular}
}
\end{table}

\paragraph{Additional Models (\texttt{GPT-OSS-120B} and \texttt{Gemini-2.5-Pro}).}
In the main paper, we restrict our scope to \texttt{GPT-5-mini} for all compared methods. In this section, we additionally evaluate \texttt{Gemini-2.5-Pro}, to more closely match the evaluation setup of ReasoningBank \citep{ouyang2025reasoningbankscalingagentselfevolving}, and \texttt{GPT-OSS-120B}, whose open weights could facilitate future white-box analysis.
We focus on WebArena with the no-memory baseline and ReasoningBank under the ordinal and Shuffle-1 task orders. We report the results in Table~\ref{tab:additional-models}. 

These results confirm two key findings of the main paper: 
(1)~Variance across runs can be large for self-improving agents (\textit{e.g.}, a 9.4-point best-worst gap on GitLab with \texttt{Gemini-2.5-Pro}, and a 13.2-point gap on Reddit with \texttt{GPT-OSS-120B}).
(2) Task order matters, and shuffled orders can push performance below the no-memory baseline (\texttt{GPT-OSS-120B}: 37.0 vs. 40.6 overall; \texttt{Gemini-2.5-Pro}: 48.0 vs. 49.1).

We also observe that variance amplification is model-dependent. ReasoningBank increases the standard deviation relative to the no-memory baseline in 8 of 12 cases for \texttt{GPT-5-mini} (main paper), but only 7 of 12 for \texttt{GPT-OSS-120B} and 5 of 12 for \texttt{Gemini-2.5-Pro}. The evidence for systematic amplification is therefore weaker with these two models. However, the broader concern remains: run-to-run variance is substantial for all three models regardless of whether memory amplifies it (\textit{e.g.}, best–worst gaps of 9.4 and 13.2 points noted above), which renders single-run evaluation unreliable.

\begin{table}[h]
\centering
\small
\caption{\textbf{Enterprise-Ops Gym Results.} Each cell reports avg/std/best-worst gap across three runs, and the right-most column (``All'') reports the overall pass rate. }
\label{tab:eog-results}
\setlength{\tabcolsep}{2pt}
\scalebox{0.85}{
\begin{tabular}{lccccccccc}
\toprule
Method & Calendar & CSM & Drive & Email & HR & Hybrid & ITSM & Teams & All \\
\midrule
\rowcolor{gray!10}\multicolumn{10}{l}{\texttt{GPT-OSS-120B}} \\
\midrule
No memory   & 21.3/2.7/6.6 & 21.4/0.8/1.9 & 41.2/2.0/4.7  & 54.2/2.5/6.0 & 12.4/0.9/2.0 & 23.6/1.5/3.7 & 5.8/0.8/1.9  & 35.5/4.3/9.8  & 24.5 \\
With memory & 26.2/2.7/6.6 & 19.7/2.8/6.8 & 33.9/6.3/14.1 & 52.7/1.4/3.0 & 5.9/1.4/2.9  & 25.6/2.0/4.9 & 1.3/0.5/1.0  & 36.1/3.5/8.2  & 22.3 \\
\midrule
\rowcolor{gray!10}\multicolumn{10}{l}{\texttt{GPT-5.4-mini}} \\
\midrule
No memory   & 38.3/0.8/1.6 & 21.4/1.6/3.9 & 47.9/2.7/6.3  & 58.7/2.5/6.0 & 20.6/1.4/2.9 & 24.8/4.0/9.8  & 25.6/3.7/7.8 & 21.3/2.3/4.9  & 30.5 \\
With memory & 36.1/1.3/3.3 & 25.9/1.7/3.9 & 51.6/3.4/7.8  & 56.7/1.2/3.0 & 16.7/2.1/4.9 & 23.6/5.8/13.4 & 24.6/1.2/2.9 & 15.3/4.7/11.5 & 29.7 \\
\bottomrule
\end{tabular}
}
\end{table}

\paragraph{Additional Benchmark (EnterpriseOps-Gym) and Baseline (Memory Filesystem).} 

The main paper focuses on web-browsing tasks and two memory-based self-improving methods. To assess the generality of our findings, we conduct a new set of experiments with an additional self-improving method and an additional benchmark.
The new method is built on the Letta Code framework \citep{MemGPT,lin2025sleeptimecomputeinferencescaling}\footnote{\faGithub\ \href{https://github.com/letta-ai/letta-code}{\texttt{letta-ai/letta-code}}}. After each task, the binary task reward is revealed. A reflection agent inspects the task trajectory and writes memory to a filesystem, where it can modify system prompts and add skills or references by creating and editing files. This design supports multi-step edits and offers greater flexibility than AWM and ReasoningBank.
The new benchmark, EnterpriseOps-Gym \citep{malay2026enterpriseopsgym}, targets tool use in enterprise settings across 8 domains (\textit{e.g.}, calendar, email, HR).
We evaluate two models (\texttt{GPT-OSS-120B} and \texttt{GPT-5.4-mini}) in this setting and report the results in Table~\ref{tab:eog-results}.

Like AWM and ReasoningBank, this third self-improving method does not consistently improve performance. It helps in some domains but hurts in others, and the overall success rate drops slightly with memory enabled (24.5$\rightarrow$22.3 for \texttt{GPT-OSS-120B}; 30.5$\rightarrow$29.7 for \texttt{GPT-5.4-mini}).
Memory also amplifies variance in the majority of cases: in 10 of 16 cases (2 models × 8 domains), the with-memory configuration shows larger variance than the no-memory baseline. Most notably, in the drive domain with \texttt{GPT-OSS-120B}, the best–worst gap increases from 4.7 to 14.1 points once memory is enabled.

\vfill
\section{Prompts}
\vspace{2cm}
\label{app:prompts}

\begin{center}
\begin{minipage}{0.95\linewidth}
\captionof{listing}{Memory construction prompt used in AWM. The prompt is obtained from the codebase of ASI \cite{wang2025inducing}.}
\label{lst:awm}
\begin{promptbox}
You are a proficient software engineer. Your task is to (1) summarize reusable workflows from the provided action trajectories. Each workflow should be a textual guideline to solve a task. 
A workflow should contain the task described in natural language, and an action trajectory to solve the task. The action trajectory should contain a sequence of steps, each with an action and a textual description. 

An example workflow look like this:
"""
Task: Tell me who has made the most contributions, in terms of number of commits, to the primer/design project
Action Trajectory:
```click("505")```  # open the `Primer / design` repository page
```click("642")```  # click on the "Commits" link
```click("307")```  # click on the "Contributors" link
```stop("The top 3 contributors to the 'prime/design' repo are: 1. Shawn Allen with 95 commits, 2. Inayaili León with 77 commits, 3. Aurora Pleguezuelo with 66 commits.")```  # This page contains the information required to determine the top 3 contributors to the `prime/design` repository, based on the number of commits. I will now extract and send the top 3 contributors based on the number of commits from the current page.

"""

Do not include the answer or example-specific information in the workflows.
Make sure each workflow has no less than 2 steps and no more than 5 steps; to keep the workflow simple and task-oriented.
You can generate zero or one workflow depending on the provided examples.

Particularly for `click` actions, either put exact element ids like `click('120')` or use a variable placeholder like `click({element_id})`. Do NOT put text strings in such as `click('Reports')`.

[Two additional example workflows were included in the experiments but skipped here.]

```
\end{promptbox}
\end{minipage}
\end{center}
\vfill

\begin{center}
\begin{minipage}{0.95\linewidth}
\captionof{listing}{Memory construction prompt used in ReasoningBank (successful tasks).}
\label{lst:rbank-successful}
\begin{promptbox}
You are an expert in web navigation. You will be given a user query, the corresponding trajectory that represents how an agent successfully accomplished the task.

## Guidelines

You need to extract and summarize useful insights in the format of memory items based on the agent's successful trajectory. The goal of summarized memory items is to be helpful and generalizable for future similar tasks.

## Important notes
- You must first think why the trajectory is successful, and then summarize the insights.
- You can extract at most 3 memory items from the trajectory.
- You must not repeat similar or overlapping items.
- Do not mention specific websites, queries, or string contents, but rather focus on the generalizable insights.

## Output Format
Your output must strictly follow the Markdown format shown below:

```
# Memory Item i
## Title: <the title of the memory item>
## Description: <one sentence summary of the memory item>
## Content: <1-3 sentences describing the insights learned to successfully accomplishing the task>
```
\end{promptbox}
\end{minipage}
\end{center}
\vfill

\begin{center}
\begin{minipage}{0.95\linewidth}
\captionof{listing}{Memory construction prompt used in ReasoningBank (failed tasks).}
\label{lst:rbank-failed}
\begin{promptbox}
You are an expert in web navigation. 
You will be given a user query, the corresponding trajectory that represents how an agent attempted to resolve the task but failed.

## Guidelines

You need to extract and summarize useful insights in the format of memory items based on the agent's failed trajectory.
The goal of summarized memory items is to be helpful and generalizable for future similar tasks.

## Important notes
- You must first reflect and think why the trajectory failed, and then summarize what lessons you have learned or strategies to prevent the failure in the future.
- You can extract at most 3 memory items from the trajectory.
- You must not repeat similar or overlapping items.
- Do not mention specific websites, queries, or string contents, but rather focus on the generalizable insights.

## Output Format

Your output must strictly follow the Markdown format shown below:
```
# Memory Item i
## Title: <the title of the memory item>
## Description: <one sentence summary of the memory item>
## Content: <1-3 sentences describing the insights learned to successfully accomplishing the task> 
```

\end{promptbox}
\end{minipage}
\end{center}

\newpage
\section{Failure Modes}
\label{app:case-studies}
\begin{table*}[h]
\caption{\textbf{Example of Ambiguous Task Query: WebArena Task 118.} The task is intended to be a web navigation task but is interpreted as a question answering task by the agent.}
\label{tab:ambiguous-task-query-example}
\centering
\small
\setlength{\tabcolsep}{8pt}
\renewcommand{\arraystretch}{1.0}
\begin{tabularx}{\textwidth}{>{\raggedright\arraybackslash}X}
\toprule
\rowcolor{gray!10}
\textbf{Task Query} \\
\midrule
\parbox[t]{\linewidth}{\raggedright
I have jaw bruxism problem, show me something that could alleviate the problem.
} \\
\midrule
\rowcolor{gray!10}
\textbf{Ground-Truth Evaluator} \\
\midrule
\parbox[t]{\linewidth}{\raggedright
``Jaw bruxism'' and ``mouth guard'' must be included in the content of the final page.
} \\
\midrule
\rowcolor{gray!10}
\textbf{Agent's Final Message} \\
\midrule
\parbox[t]{\linewidth}{\raggedright
\textit{(The agent didn't browse the shopping website and ends the task after 1 step.)}\\
Here are practical things that can help reduce jaw bruxism (teeth grinding/clenching). This is general information --- consult your dentist or doctor for personalized advice:\\
1) Night guard / oral splint\\
- Best: a custom-fitted occlusal guard from your dentist (protects teeth and reduces muscle stress). [text omitted for brevity...]
} \\
\midrule
\rowcolor{gray!10}
\textbf{Agent's Memory} \\
\midrule
\parbox[t]{\linewidth}{\raggedright
\textbf{Title:} Always gather targeted patient/context details before giving medical guidance\\
\textbf{Description:} Before offering specific treatments for a health problem, ask concise clarifying questions about symptoms and context.\\
\textbf{Content:} Ask whether the problem occurs during sleep or wakefulness, severity/frequency, current treatments, dental history, relevant medications, pregnancy, and presence of red-flag symptoms (severe pain, tooth damage, jaw locking, daytime sleepiness). These details let you tailor recommendations (e.g., conservative self-care vs.\ urgent dental referral) and avoid unsafe suggestions.
} \\
\bottomrule
\end{tabularx}
\end{table*}

\vspace{0.2cm}
\begin{table*}[h]
\caption{\textbf{Example of Evaluator Bug: WebArena Task 329.} A correct answer may fail strict string-matching rubric, and further lead to the creations of irrelevant memories.}
\label{tab:evaluator-bug-example}
\centering
\small
\setlength{\tabcolsep}{8pt}
\renewcommand{\arraystretch}{1.0}
\begin{tabularx}{\textwidth}{>{\raggedright\arraybackslash}X}
\toprule
\rowcolor{gray!10}
\textbf{Task Query} \\
\midrule
\parbox[t]{\linewidth}{\raggedright
How much I spend on 4/19/2023 on shopping at One Stop Market?
} \\
\midrule
\rowcolor{gray!10}
\textbf{Ground-Truth Evaluator} \\
\midrule
\parbox[t]{\linewidth}{\raggedright
The answer string must include ``0''.
} \\
\midrule
\rowcolor{gray!10}
\textbf{Agent's Final Message} \\
\midrule
\parbox[t]{\linewidth}{\raggedright
I could not find any orders dated 4/19/2023 in your One Stop Market account. I scanned the full order history (37 orders shown after setting per-page to 50) and the extraction returned no matches for 4/19/2023. Therefore, according to this account, the amount spent on 4/19/2023 at One Stop Market is \$0.00.
} \\
\midrule
\rowcolor{cyan!10}
\textbf{Author Comment} \\
\midrule
\parbox[t]{\linewidth}{\raggedright
``0.00'' is processed as one token and is considered different from ``0'' according to WebArena evaluator.
} \\
\midrule
\rowcolor{gray!10}
\textbf{Agent's Memory} \\
\midrule
\parbox[t]{\linewidth}{\raggedright
\textbf{Title:} Consider alternate accounts and guest/third-party purchases\\
\textbf{Description:} Missing orders in one account can result from purchases made under a different account, guest checkout, or through a different platform.\\
\textbf{Content:} If an order date isn't found, check other user accounts, alternate email addresses or phone numbers, loyalty profiles, and third-party marketplaces or apps the user might have used.
} \\
\bottomrule
\end{tabularx}
\end{table*}

\begin{table*}[h]
\caption{\textbf{Example of Using Haversine Formula: WebArena Task 38.} The agent receives an reward for including ``yes'' in its final answer, while the strategy to reach the answer is unintended.}
\label{tab:duquesne-one-hour-eval}
\centering
\small
\setlength{\tabcolsep}{8pt}
\renewcommand{\arraystretch}{1.0}
\begin{tabularx}{\textwidth}{>{\raggedright\arraybackslash}X}
\toprule
\rowcolor{gray!10}
\textbf{Task} \\
\midrule
\parbox[t]{\linewidth}{\raggedright
Check if Duquesne University in Pittsburgh can be reached in one hour by car from Pittsburgh Airport.
} \\
\midrule
\rowcolor{gray!10}
\textbf{Rubric} \\
\midrule
Answer string must include "yes".\\
\midrule
\rowcolor{gray!10}
\textbf{Agent's Answer} \\
\midrule
\textbf{Result and conclusion:}\\
\textit{What I could not do:} The OpenStreetMap directions page did not show a numeric route or duration because it displayed messages indicating JavaScript was not available, so a live OSRM route summary could not be extracted.\\
\textit{Fallback method used:} Coordinates embedded in the page URL were used (origin: 40.4918, -80.2370 --- Pittsburgh International Airport; destination: 40.4365, -79.9905 --- Duquesne University area). Great-circle (haversine) distance was computed, then a conservative routing factor was applied to estimate driving distance and time.\\
\textit{Calculations and assumptions:}\\
\textbullet\ Straight-line distance (haversine): \(\approx 21.7\) km (\(\approx 13.5\) miles).\\
\textbullet\ Estimated driving distance (routing factor \(\approx 1.3\)): \(\approx 28\)--29 km (\(\approx 17.5\)--18 miles).\\
\textbullet\ Estimated driving time using typical effective speeds: 60 mph \(\rightarrow\) \(\sim 17\)--18 min; 40 mph \(\rightarrow\) \(\sim 26\)--28 min; heavy congestion at \(\approx 30\) mph \(\rightarrow\) \(\sim 35\)--40 min.\\
\textbullet\ Conservative upper bound: under most circumstances, the drive remains well under 60 minutes; only extreme incidents may push it beyond one hour.\\
\colorbox{green!20}{\textit{Final answer:} Yes}
 --- Duquesne University can be reached by car from Pittsburgh International Airport in under one hour under normal and moderately heavy traffic conditions (typical estimate: \(\sim 20\)--40 minutes). For exact current travel time, enable JavaScript on the directions page or use a live routing API (OSRM/Google Maps/Waze).\\
\textit{Method summary (reproducibility):} Used URL coordinates (40.4918,-80.2370 \(\rightarrow\) 40.4365,-79.9905), computed haversine distance (\(\approx 21.7\) km), applied a 1.3 routing multiplier (\(\approx 28\)--29 km), and converted distance to time ranges using reasonable average speeds.
\\\midrule
\rowcolor{gray!10}
\textbf{Agent's Memory} \\
\midrule
\textbf{Title:} Use URL or page coordinates as a robust routing fallback.\\
\textbf{Description:} If the directions UI or routing engine fails to render, extract coordinates from the page or URL and compute an offline route estimate.\\
\textbf{Content:} Parse origin and destination latitudes and longitudes embedded in the page or URL, compute straight-line (haversine) distance, and convert it to plausible driving distance using a routing multiplier (e.g., \(\sim 1.2\)--1.4). This provides a reliable fallback when live routing is unavailable. \\
\bottomrule
\end{tabularx}
\end{table*}

\newpage


\clearpage
\section*{NeurIPS Paper Checklist}

\begin{enumerate}

\item {\bf Claims}
    \item[] Question: Do the main claims made in the abstract and introduction accurately reflect the paper's contributions and scope?
    \item[] Answer: \answerYes{} 
    \item[] Justification: Our abstract and introduction reflect our contributions and scope.

\item {\bf Limitations}
    \item[] Question: Does the paper discuss the limitations of the work performed by the authors?
    \item[] Answer: \answerYes{} 
    \item[] Justification: \S\ref{app:limitations}

\item {\bf Theory assumptions and proofs}
    \item[] Question: For each theoretical result, does the paper provide the full set of assumptions and a complete (and correct) proof?
    \item[] Answer: \answerNA{} 
    \item[] Justification: Our work does not involve theoretical analysis.

    \item {\bf Experimental result reproducibility}
    \item[] Question: Does the paper fully disclose all the information needed to reproduce the main experimental results of the paper to the extent that it affects the main claims and/or conclusions of the paper (regardless of whether the code and data are provided or not)?
    \item[] Answer: \answerYes{} 
    \item[] Justification: We discuss experiment settings and details in \S\ref{ssec:experiments} and \S\ref{app:additional-experiment-details}.

\item {\bf Open access to data and code}
    \item[] Question: Does the paper provide open access to the data and code, with sufficient instructions to faithfully reproduce the main experimental results, as described in supplemental material?
    \item[] Answer: \answerYes{} 
    \item[] Justification: Released at \faGithub\ \href{https://github.com/SalesforceAIResearch/self-improve-fragility}{\texttt{SalesforceAIResearch/self-improve-fragility}} and \faHuggingFace\ \href{https://huggingface.co/datasets/Salesforce/self-improve-fragility}{\texttt{Salesforce/self-improve-fragility}}

\item {\bf Experimental setting/details}
    \item[] Question: Does the paper specify all the training and test details (e.g., data splits, hyperparameters, how they were chosen, type of optimizer) necessary to understand the results?
    \item[] Answer: \answerYes{} 
    \item[] Justification: We describe the dataset we used and discuss experiment details in \S\ref{ssec:experiments} and \S\ref{app:additional-experiment-details}.

\item {\bf Experiment statistical significance}
    \item[] Question: Does the paper report error bars suitably and correctly defined or other appropriate information about the statistical significance of the experiments?
    \item[] Answer: \answerYes{} 
    \item[] Justification: We include error bars whenever possible (\textit{e.g.}, Fig.~\ref{fig:task-order-result-overview}, Fig.~\ref{fig:additional-info-for-memory-construction}). In fact, one main goal of this paper is to encourage reporting multi-run statistics responsibly.

\item {\bf Experiments compute resources}
    \item[] Question: For each experiment, does the paper provide sufficient information on the computer resources (type of compute workers, memory, time of execution) needed to reproduce the experiments?
    \item[] Answer: \answerYes{} 
    \item[] Justification: See \S\ref{app:additional-experiment-details}.
    
\item {\bf Code of ethics}
    \item[] Question: Does the research conducted in the paper conform, in every respect, with the NeurIPS Code of Ethics \url{https://neurips.cc/public/EthicsGuidelines}?
    \item[] Answer: \answerYes{} 
    \item[] Justification: The paper conform with the NeurIPS Code of Ethics.

\item {\bf Broader impacts}
    \item[] Question: Does the paper discuss both potential positive societal impacts and negative societal impacts of the work performed?
    \item[] Answer: \answerYes{} 
    \item[] Justification: We discuss implications of our work and practical recommendations stemming from our findings in the conclusion section (\S\ref{sec:conclusion}).
    
\item {\bf Safeguards}
    \item[] Question: Does the paper describe safeguards that have been put in place for responsible release of data or models that have a high risk for misuse (e.g., pre-trained language models, image generators, or scraped datasets)?
    \item[] Answer: \answerNA{} 
    \item[] Justification: This paper does not introduce new datasets or models. While we release the agent trajectories collected in this study, we do not anticipate a significant risk of misuse.

\item {\bf Licenses for existing assets}
    \item[] Question: Are the creators or original owners of assets (e.g., code, data, models), used in the paper, properly credited and are the license and terms of use explicitly mentioned and properly respected?
    \item[] Answer: \answerYes{} 
    \item[] Justification: See \S\ref{app:additional-experiment-details}.

\item {\bf New assets}
    \item[] Question: Are new assets introduced in the paper well documented and is the documentation provided alongside the assets?
    \item[] Answer: \answerYes{} 
    \item[] Justification: Our code and data are released under Apache-2.0 license.

\item {\bf Crowdsourcing and research with human subjects}
    \item[] Question: For crowdsourcing experiments and research with human subjects, does the paper include the full text of instructions given to participants and screenshots, if applicable, as well as details about compensation (if any)? 
    \item[] Answer: \answerNA{} 
    \item[] Justification: This paper does not involve crowdsourcing or research with human subjects.

\item {\bf Institutional review board (IRB) approvals or equivalent for research with human subjects}
    \item[] Question: Does the paper describe potential risks incurred by study participants, whether such risks were disclosed to the subjects, and whether Institutional Review Board (IRB) approvals (or an equivalent approval/review based on the requirements of your country or institution) were obtained?
    \item[] Answer: \answerNA{} 
    \item[] Justification: We did not recruit study participants in this work and hence this is not applicable.

\item {\bf Declaration of LLM usage}
    \item[] Question: Does the paper describe the usage of LLMs if it is an important, original, or non-standard component of the core methods in this research? Note that if the LLM is used only for writing, editing, or formatting purposes and does \emph{not} impact the core methodology, scientific rigor, or originality of the research, declaration is not required.
    \item[] Answer: \answerYes{} 
    \item[] Justification: See \S\ref{app:additional-experiment-details}.

\end{enumerate}

\end{document}